# Responsible AI Implementation: A Human-centered Framework for Accelerating the Innovation Process

Dian Tjondronegoro*
*Griffith University, Brisbane, Australia*
Elizabeth Yuwono
*Griffith University, Brisbane, Australia*
Brent Richards
*Gold Coast University Hospital, Gold Coast, Australia*
Damian *Green*
*Queensland Health, Brisbane, Australia*
Siiri Hatakka
*Everledger, Brisbane, Australia*
* Contact author: d.tjondronegoro@griffith.edu.au

**Abstract**
There is still a significant gap between expectations and the successful adoption of AI to innovate and improve businesses. Due to the emergence of deep learning, AI adoption is more complex as it often incorporates big data and the internet of things (IoT), affecting data privacy. Existing frameworks have identified the need to focus on human-centered design, combining technical and business/organizational perspectives. However, trust remains a critical issue that needs to be designed from the beginning. The proposed framework is the first to expand from the human-centered design approach, emphasizing and maintaining the trust that underpins the whole process. This paper proposes a new theoretical framework for responsible artificial intelligence (AI) implementation. The proposed framework emphasizes a synergistic business-technology approach for the agile co-creation process. The aim is to streamline the adoption process of AI to innovate and improve business by involving all stakeholders throughout the project so that the AI technology is designed, developed, and deployed in conjunction with people and not in isolation. The framework presents a fresh viewpoint on responsible AI implementation based on analytical literature review, conceptual framework design, and practitioners' mediating expertise. The framework emphasizes establishing and maintaining trust throughout the human-centered design and agile development of AI. This human-centered approach is aligned with and enabled by the "privacy-by-design" principle. The creators of the technology and the end-users are working together to tailor the AI solution specifically for the business requirements and human characteristics. An illustrative case study on adopting AI for assisting planning in a hospital will demonstrate that the proposed framework applies to real-life applications.

## 1. Introduction

Despite the rapid advancement and growth of global investment in Artificial Intelligence (AI) technologies, a survey with more than 3000 business executives has recently revealed that only a fifth have incorporated AI in their processes (Ransbotham et al. 2017). The lack of adoption shows a significant gap between ambition and execution for implementing AI to innovate and improve businesses. Meanwhile, due to the rapid growth of AI-enabled automation, people become more worried about losing jobs and control over their data. Businesses pursue innovation to maintain their competitive edge, develop new market segments, and improve the quality of their products and organizational practices. Technology is often a vital part of the catalyst toward more opportunities and productivity. However, we must work on the people, organization, leadership, governance, policy, and strategy to make digital transformations happen.

AI adoption as a form of digital innovation is at the intersection between viability (business), feasibility (technology), and desirability or usability (human factor). A successful digital transformation will lead to empowered employees, engaged customers, transformed products, and optimized operations. Current frameworks for designing, developing, and deploying enterprise technologies have only implicitly shown the connections between technical design and business adoption strategies. However, there is yet a framework that emphasizes the integration of technology and business throughout the process. The integration requires leadership for the change process and a concrete framework to enable the implementation process to bring about successful AI integration into practice.

One major hurdle for AI adoption is managing scope and expectation while keeping trust in the technology. Most people and organizations still do not have a full understanding of AI and its associated technologies. Trust and confidence in AI start with a 'problem worth solving' and evidence that the initial AI solution can prove the feasibility of using AI to solve the problem. Once the value of the AI solution is clear, people and organizations are more likely to accept the proposed business model, changes to workflow, and the continuing adaptation of the AI solution. Along with the trust and confidence in the technology, the organizational maturity level needs to grow in its readiness to innovate with AI (Sicular et al., 2020). At the earliest (planning) stage, pioneers must explore available AI models and prioritized business use cases. Next is the experimentation stage to prove the business value of AI use cases, leading to the stabilization of infrastructure (people, process, and technology) to support data access and AI tools. Ongoing AI strategy needs to be owned by the executives to ensure that the AI innovation is ready to scale and expand. At the most mature (transformation) stage, the business integrates AI seamlessly into the process and shares the benefits with the public by delivering AI-enabled outcomes, including new products and services.

Most of the AI-based solutions require ongoing machine learning to use data and improve the models and performance continually. Therefore, trust formation in the AI's initial model needs to be followed up with continuous maintenance of the trust while progressing on the AI application's design and development (Siau and Wang, 2018). This paper will present a framework to enable responsible AI implementation that incorporates all stakeholders' inputs throughout the co-creation process to realize the identified shared value and objectives fully. The framework presents a fresh viewpoint on responsible AI implementation based on analytical literature review, conceptual framework design, and practitioners' mediating expertise. The framework emphasizes establishing and maintaining trust throughout the human-centered design and agile development of AI.

This paper aims to support academics and practitioners from across business and technology in their innovation journey to implement AI in real-life applications. Section 2 will discuss AI in the context of the Internet of Things and big data to outline the importance of trust, privacy, and security for responsible computing. Section 3 then will describe the requirements for responsible AI and discuss the gaps in existing frameworks in the literature from both academics and practitioners. Section 4 will discuss the proposed framework in detail, followed by an illustrative case study (in Section 5) on using AI for assisting planning in hospitals to demonstrate the applicability of the proposed framework for real-life applications.

## 2. AI in the era of IoT and Big Data

AI is the computer's ability to 1) process data into useful information, 2) analyze extracted information to support decision making and augment human's knowledge, 3) develop cognitive capabilities, which would enable machines to ultimately 4) become autonomous for the task presented and act according to logic, context, rules, and laws. AI is the central information processing capability that enables computers to think and act. Just like how a human's brain grows its intelligence and ability to make decisions, AI needs to be trained by big data, connected to the source of knowledge produced by the Internet of Things and governed by laws and rules to assist and take responsible actions and decisions.

To fully appreciate why AI implementation is a complex issue, this section provides a high-level summary of the technical aspect of AI and its interconnecting technologies. This section aims to help readers understand the requirements for responsible AI and its potential implications.

*AI and Big Data: how they are symbiotically dependent and benefit each other*

There are two methods to train a machine's intelligence. First is by programmatically teaching it with established rules and structures based on human experts. The second is by letting it learn by itself based on deep learning algorithms without human intervention. Deep learning enables AI to autonomously extract useful features from any data and identify patterns to continuously improve its knowledge and decision-making ability (LeCun *et al*., 2015). The application of deep learning algorithms has triggered the continuous and rapid development of AI implementations, from helping a computer perform cognitive tasks and making complex decisions (Benbya and McKelvey, 2006). Deep learning can deliver beyond human intelligence in specific tasks, such as detecting objects from millions of images within a few hours while rapidly learning new object classes based on data (see Vernon *et al.* 2007 for a survey on AI cognitive abilities).

Deep learning-based AI requires many samples to develop logic, knowledge, perception, and cognitive capabilities. If we think about how medical doctors are trained based on accumulated and particular cases of previous experiences, AI also requires accumulated data to establish a capacity to make decisions based on the current situation that has to be carefully perceived and analyzed. The more data AI can leverage, the more accurate and useful it can become. Big data signifies the requirement and the capacity of today's computers to produce and process data and information at high volume, velocity, variety, and veracity. Therefore, big data is like the oil to an AI engine, as without it, AI would not be able to run accurately. On the other hand, without AI, big data would continue to be underutilized and ultimately discarded. It is generally cumbersome and humanly impossible to rapidly make sense of and extract meaningful information from a massive amount of unstructured data continuously and in a short time. Therefore, AI turns big data into an asset.

*The Internet of Things and Cybersecurity: how they affect AI considerations*

The Internet of Things (IoT) is the interconnected network of computing devices embedded in everyday objects so that they can communicate via data with or without human interventions. When everything in the world is connected, we will witness a sharp increase in data volume generated by users' daily life experiences, businesses' transactional records, connected devices, and social interactions. Big data generated by IoT can be scaled and managed effortlessly using cloud services, incorporating the ever-increasing number of new devices and data types. However, these raw data cannot translate itself into useful knowledge, which is where AI would play a crucial role in interpreting data into useful information and knowledge. As millions of connected things are generating big data, it will support AI's deep learning. IoT-generated big data is crucial to computationally reveal useful patterns, trends, and associations to learn and interpret behavior and interactions between humans, machines, and the environment.

Figure 1 describes how intelligent systems collect data from the environment through the IoT. Sensors networks monitor people and (living) things within various environments, producing big multimedia data that feeds into intelligent systems for interpretation and visualization. The intelligent systems can enable intuitive and useful pervasive UIs to help augment human intellect to solve grand research challenges and daily activities. When AI is used to analyze data generated from millions of interconnected devices to take actions or make decisions, the quality and integrity of the data must be of the highest standard. Cybersecurity in the IoT era is even more prominent and challenging issue to prevent data theft and tampering. A primary challenge is ensuring that each device's safety and privacy features and data transmission are sufficient. The people and organization are enforcing behavior and code of practice that promotes Internet security.

Ethical use of data and AI is ever more important as most individuals, businesses, and services have started to rely on a vast amount of knowledge and information to make better decisions and anticipate new challenges. People nowadays start to give away their privacy in return for convenience and tailored services based on their preferences and contextual requirements. This phenomenon will continue to change the future governance and law of private and shareable information, determining the availability of data for AI's machine learning.

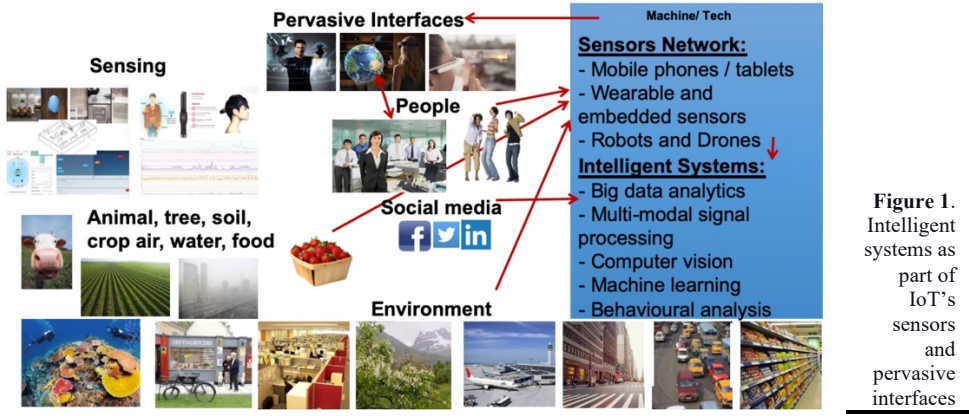

**Figure 1**. Intelligent systems as part of IoT's sensors and pervasive interfaces

### 3. Requirements for Responsible AI
Responsible AI signifies the development of intelligent systems that maintain fundamental human principles and values to ensure human flourishing and well-being in a sustainable world (Dignum, 2019). Taking privacy, ethics, and fairness issues into consideration, the business case for AI adoption is challenging as it requires mastery of different technologies and problem-solving skills. Moreover, adopting AI into an organization often requires confidential data and a reliable tool to explain AI as a 'black box.' Most users cannot understand the complexity of advanced AI's internal workings (Hajkowicz *et al.*, 2019) – see (Samek and Muller, 2019) for detail on explainable AI requirements.

Table I summarizes the principles of responsible Robotics (Deloitte, 2019), which comprise security, safety, privacy, fairness, sustainability, accountability, and transparency. Given the large scope of responsible robotics, AI adoption will require a substantial cultural change and transition to new work roles and practices. The shift requires leadership, a change of thinking and behavior in the organization, and an open dialogue across all the stakeholders to gain mutual understanding and respect. There is also a need to spend more time collecting data for training the AI models and algorithms, which largely depend on the existing data and analytics infrastructure. Moreover, it can be challenging to predict the AI decision due to dynamic data and heuristic solutions rather than fixed rules. As such, there is a need to maintain transparency and openness in all decisions, primarily when the organization adopts a combination of in-house development and outsourcing for implementing AI (Andrews, 2018).

*Requirements of Responsible AI Implementation in Businesses*
Responsible AI adoption in organizations emphasizes creating systems that are ethical, understandable, and legal and incorporate a test for bias in data, models, and human use of algorithms. (PwC, 2019). We need the appropriate ethics and regulation to demonstrate ongoing governance on the performance of AI's capacity. The main issues are bias and fairness, interpretability or explainability, and robustness against security threats.

People's trust in the technology and the company (that manages it) is essential to achieve widespread AI adoption to support decision-making and problem-solving. Trust amid perceived risk can be achieved from overall transparency at both the technology and company levels (Gasson and Shelfer, 2007). At the technology level, the AI solution needs to provide compelling user interfaces and comprehensive product manuals, scientific data, technical explanations, scientific reviews, and technical testing. At the company level, fundamental practice for transparency in the leadership, management, and documentation is critical, including making all trials visible and fully acknowledged. Therefore, implementing governance for responsible AI adoption requires organizational leadership to cultivate transparent communication and a culture of responsibility and align the internal processes with regulations and industry best practices. The governance needs to be embedded within the daily practices and processes of the project teams, focusing on engaging and educating the customers to provide consistent transparency.

| | |
|---|---|
| Security | - Technology architecture is rigorously tested for vulnerabilities.<br>- The protective measure is tested periodically verifying safe and protected from security threats. |
| Safety | - Operationally safe cannot jeopardize the physical safety and wellbeing of humanity.<br>- Solutions are sufficiently tested and have clear usage instructions. |
| Privacy | - The solution aligns with privacy by design practice.<br>- Personal data is processed lawfully to a minimum, and data is kept safe from internal or external influences. |
| Fairness | - The solution does not operate in a discriminative way, protecting fundamental rights, and ethical principles.<br>- The purpose and tasks of the solution are clearly stated and designed for the full range of human abilities. |
| Sustainability | - The solution aligns with fair employment and labor practices, taking into account social and cultural justice.<br>- Actions are taken to reduce the negative environmental impact. |
| Accountability | - The solution complies with laws, fundamental rights, and freedoms.<br>- Actions, intent, and decisions are traceable. |
| Transparency | - Transparent about the use of an AI engine.<br>- Decisions made are explainable, traceable, and auditable. |

**Table I.** Seven principles for responsible robotics (summarized from Deloitte, 2019)

Human-machine synergistic efforts can ultimately ensure the safety, fairness, and auditability of the continuous improvements of machine learning algorithms and real-time data to achieve better outcomes. The AI models need to reliably continue functioning as intended after deployment into production, which requires active monitoring of:
- user-behavior to prevent unintended consequences,
- performance drift management to monitor the model's performance and spot requirements to initiate retraining,
- operational bias to identify irregularities and biases in the outputs, and
- model retraining to use new data to address the changes.

(Delmolino and Whitehouse, 2018).

AI relies on sufficiently representative and diverse data to effectively train the models for making reliable and unbiased decisions. *Transparent* and *Explainable AI* aims to apply new processes, technologies, and layers to existing AI systems to make AI models understandable to users and programmers. The human-in-the-loop concept denotes human involvement in the design and evaluation processes to maintain a supervisory role over autonomous systems. The supervisory process ensures that AI behaves in alignment with human rights, social norms, and privacy practices and oversees automation biases.

*Privacy, Ethics, and Fairness throughout AI implementation*
Privacy and processing of personal data are an essential part of an individual's freedom and fundamental rights as well as society's democracy. Some academics and practitioners currently consider Europe's General Data Protection Regulation (GDPR) as the most cutting-edge world regulation on personal data protection (Inverardi, 2019). GDPR is the first legislation to emphasize "privacy by design" as the critical acceptance criteria of AI-based systems, which may become an enabler for maintaining trust but also restrictive due to excess barriers to implementation. Given the scope and amount of data used by AI, ethical privacy practices need to be embedded into the design specifications of technologies, business practices, and physical infrastructures. A growing number of emerging privacy-preserving measures must be applied to the hardware, software, and datasets.

Beyond privacy, ethical issues are the hardest to manage due to the potential conflict between users' and communities' ethical principles for an autonomous system's decisions. The challenge is to balance the freedom of individuals to make responsible decisions and the freedom and safety of others. For example, in the AI-enabled self-driving context, individuals should be able to maintain their freedom to make decisions. Therefore, the challenge is whether AI should ethically support individuals to exercise their freedom of choice or enforce the law by autonomously intervening in the decisions and taking control of the vehicle for the sake of others' benefits and safety. Another challenge is whether AI should be ethically required to provide all data and information for law enforcement officers to make a fair decision or whether AI should still maintain an individual's privacy rights and wait for consent before disclosing. GDPR article 22 states that human beings have the right to be not solely subjected to automated decisions when such decisions have legal ramifications. As AI systems have no moral authority, they cannot be held accountable for judicial decisions and judgments (Dawson *et al*., 2019).

Fairness is another major issue since the data fed into its machine learning algorithms largely determines AI's ability to analyze information and make decisions. An issue is that the data may be biased, reflecting the AI's algorithms and outcomes. The rapid adoption of deep learning has shifted the development of AI capability from intellectual engineering to data-driven engineering. Therefore, there is a general expectation that AI is biased for particular contexts where the data is collected. For example, image-based facial expression recognition is still broadly trained based on unbalanced datasets, thereby causing the recognition to be influenced by racial bias (Rhue, 2018).

*Existing Gaps in Frameworks for Responsible AI Implementation*
As a form of digital transformation, AI involves people, technology, and management (Ionology, n.d.). On the people side, staff and end-user engagement must go through the innovation and embedding of technology into processes. On the technology side, it is vital that the right technology is embraced, as it is rarely the technology that enables digital transformation. On the management side, the right digital strategy and culture can drive the digitalization of services.

There is a standard framework for enterprise architecture (Open Group, 2018) to implement AI as a form of enabling technology for business. The framework highlights that understanding business operations will help to determine the business case for AI. Once the need (for using AI) is confirmed, the requirements management will need to focus on migration planning, implementation governance, and architecture change management. A key lesson from this framework is designing, developing, and deploying enterprise technologies to combine technical design and business adoption strategies.

AI maturity of the business and its industry sector will determine the journey of implementing an AI solution. Business leaders are still wary of generally "inflated" business cases for AI, as most modern AI and machine learning algorithms and models are still relatively new and not mainstream. It is important to establish concrete short-term evidence of successful applications to reduce the risk of overpromising AI's benefits. Through a rapid iterative cycle of design, development, and deployment of AI solutions, "beta" AI solution(s) can be delivered for the end-users to confirm the benefits. Such an approach also inherently reduces the risk of over-investment in time, resources, and budget for designing, developing, and maintaining the AI solution (Marcinkowski and Gawin, 2018; Saunila *et al*. 2019).

Existing frameworks presented in most recent academic papers have focused on either the general policy of responsible AI or the security perspective for IoT and AI integration (Floridi *et al*. 2018; Wirtz and Muller, 2019). The applications include IoT-enabled AI systems for the smart government (Kankanhalli *et al*. 2019) and smart and secure IoT and AI integration frameworks for the hospital environment (Valanarasu, 2019). These existing frameworks have emphasized the importance of trust and collaboration for AI deployment. The ethical use of AI enables organizations to achieve socially acceptable use of AI, which subsequently helps minimize costly mistakes from socially unacceptable or legally questionable practices. Ethics must work within an environment of public trust and clear responsibilities from all the stakeholders, which can be achieved by involving developers, users, and rule-makers to collaborate from the outset.

Many existing practitioner-led frameworks can help to guide the process of corporate AI implementation. For example, Accenture's framework for building trust in AI solutions (Delmolino and Whitehouse, 2018), PwC's AI adoption roadmap (PwC, 2019), Gartner's CIO guide to building the strategy and business case to implement AI in the enterprise (Pettey, 2018), and McKinsey's guide for leading organization to responsible AI (Burkhardt *et al.* 2019). Across all these frameworks, the fundamental emphasis is the need to develop the corporate strategy and planning for portfolio management. The next step is to design a delivery approach and program oversight and prepare for the ecosystem (technology roadmap and change management) before developing and deploying the solution iteratively, supported by operational support and compliance. The final stage is operating and monitoring the solution. This framework needs to be iterative and aligned with the human-centered design framework, which emphasizes the continual development of technology to be eventually deployed and scaled. People and organizations need to work together throughout the process of planning, designing, developing, and evaluating prototypes for business and technology to be better aligned and integrated.

AI solutions need strong data privacy protection to maintain users' trust, whereby personal and corporate information needs to be consistently accurate and protected. Business leaders must ensure that AI implementation starts with a strong business case and a clear strategy for aligning people, processes, and technology. Therefore, there is a need to extend the existing business AI adoption frameworks to address the responsible AI requirements. A crucial gap is a focus on trust formation and maintenance of trust throughout the technology development and refinement. Trust is a critical requirement, given that AI deployment requires staff and customers' support throughout the process (Qi and Chau, 2013). Hence, we need to focus on people from the start to the end to maintain a sense of shared purpose and benefits.

## 4. Proposed Framework for Responsible AI Implementation

The proposed Responsible AI Implementation Framework (RAIIF – see Figure 2) extends the human-centered design framework with an iterative co-creation process to achieve a better synergy between technology development and business integration, underpinned by trust establishment and maintenance. The framework aims to streamline the AI adoption process by involving all stakeholders throughout the implementation process, from planning and design to development and scaling up. An iterative and agile approach applies to the whole journey of planning, design, development, deployment, operating, monitoring, and scaling the AI solution. Each process may be iterated on its own or after the whole journey (from planning to scaling) of the initial AI solution is completed. Every business needs to go through the iterations during each of their own AI maturity levels - from planning, experimentation, stabilization, expansion, to transformation (Sicular *et al.* 2020). Before a business can start sharing AI advantages with others, it needs to establish the business case, stabilize the people, process, and technology infrastructure to access data and AI tools, and expand the data sources to scale up the AI models.

RAIIF is inspired by GDPR's privacy-by-design principle, with both the creators of the technology and the end-users working together to tailor the AI solution specifically for the business requirements and human characteristics. The scope and use of data to build and maintain the AI models are co-designed from the beginning to ensure a responsible and ethical practice of data management. The initial trust is formed by designing for trustability, which can be gained by ensuring the quality of information, systems, and interactions (Nasirian *et al*. 2017). Trust is continuously preserved through the iterative AI solution development by maintaining and improving the model's consistency and explainability. It also ensures that the AI's purpose of supporting human tasks and intelligence is governed by expectations and limitations on accuracy and bias.

Given that the AI algorithms and datasets used may change over time, the accuracy and, therefore, its utility may also be affected. The design and maintenance of AI solutions need to accommodate a continuous process of initial trust formation and continuous trust development. However, the more "AI-mature" a business is, the more the process can be streamlined and a full cycle of planning and designing the ecosystem process may no longer be needed. Business AI-readiness refers to the organization's people, processes, and technological capacity to effectively use AI technology.

*Co-creation across the process*
Responsible AI implementation needs all stakeholders to have a distributed responsibility in conjunction with knowledge about long-term societal and technological consequences. The framework advocates for a co-creation process, signifying that all stakeholders must be involved and committed to responsible AI implementation. All stakeholders must work together from the beginning to the end, from initial business planning to design, development, and deployment. Stakeholders are everyone involved in technical development, operational process, business model, and decision making, as well as end-user, communities, and policymakers.

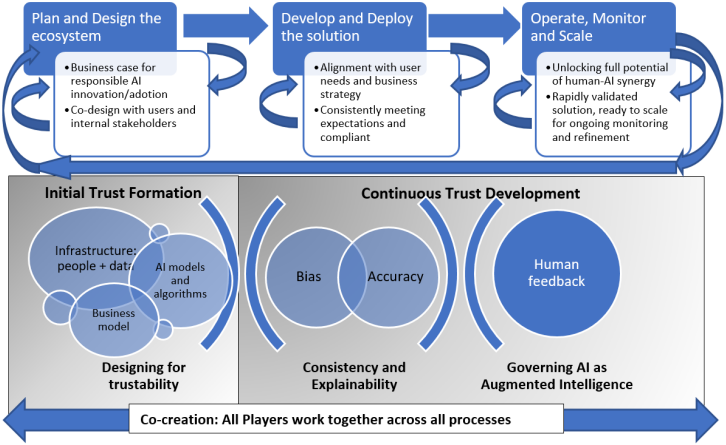

**Figure 2.** Proposed Framework for Responsible AI Implementation – this framework is iterative across all co-creation processes (planning, design, and development)

Every stakeholder shares the responsibility to uphold the three categories of responsible AI requirements, as outlined in Table II: risk management, human control, and quality control (Clarke, 2019). For example, technical developers cannot afford to have a lack of accountability for the moral significance of their work. Business decision-makers cannot allow economic incentives to override the commitment to ethical principles and values (Hagendorf, 2020). As summarized in Table III, each stakeholder needs to participate in ensuring specific requirements for responsible AI. For example, business leaders and executive decision-makers must ensure full support for investing the time, process, and technology.

| | | |
|---|---|---|
| Risk management | • Assess positive and negative impacts and implications. <br> • Ensure accountability and obligations. <br> • Enforce liability and sanctions. | |
| Human control | • Complement humans. <br> • Ensure human control. <br> • Ensure human safety and wellbeing. <br> • Ensure consistency with human values and rights. | **Table II.** Three categories of responsible AI requirements (summarized from Clarke, 2019) |
| Quality control | • Embed quality assurance. <br> • Exhibit robustness and resilience. | |
| Technical development and business operational teams | • A synergistic approach to design, develop, train, test, and continually improve the intelligent systems and the supporting human and digital resources/infrastructure.<br>• The AI solution can effectively support business activities. | |
| Business leaders and executive decision-makers | • Full support for investing time, people, process, and technology to deliver<br>• Agreed expected return of investment (ROI)<br>• AI implementation is synergistic with the business strategy, planning, leadership, management, and operational governance. | **Table III.** Requirements for distributed responsibility in AI Implementation |
| End users | • Alignment with expectation and needs.<br>• AI decisions can be explained, understood, and justifiable. | |
| Community and policymakers | • Compliance with standards, policy, ethics, governance, best practice, values, law for privacy, and security. | |

*Plan and Design the ecosystem*
The goal of the design and planning process is to achieve AI trustability, which emphasizes technological accessibility and AI-enabled business refinement, customer understanding and satisfaction, and operations monitoring. It also involves considerations and planning for AI adoption's impact on the existing human and technology infrastructure. The business model needs to consider the requirements for the ecosystem of digital and human infrastructure and the initial AI model to determine the business viability, technological feasibility, and human desire. Table IV outlines the key considerations and checklist of this process, which can serve as the critical criteria for completing it.

*Development and Deployment*
The goal of development and deployment is to ensure that the AI solution is consistently accurate, meeting expectations, and compliant. The underlying AI model must be explainable regarding how the decisions were made, as machine learning can be biased towards the specific dataset and its application contexts. The goal of the development and deployment process is to achieve AI *consistency and explainability* until it reaches the point where the AI solution can be scaled up into the full production stage and ultimately transform the end-users and society. Table V outlines the key considerations and checklist for this process, which can serve as the critical criteria for completing it.

*Operate, Monitor, and Scale-up*
The aim of operating, monitoring, and scaling AI solution is to achieve optimized operations, empowered employees, engaged customers, and transformed products. While AI can automate decisions and actions, it should augment intelligence, which means that human-machine intelligence works synergistically, particularly for critical decisions. Human in the loop is also crucial for rapidly validating the evolution of AI solution by monitoring the results and refining them based on manual feedback. Hence, the goal of the operate, monitor, and scale process is to unlock the full potential of AI with *human feedback* via continuous refinement and transparent business practice, particularly on the use of data and AI. This goal will ensure that the AI solution can harmoniously co-exist with the businesses and community while maintaining individual and societal trust in the technology and company amid perceived risks. Table VI outlines the key considerations and checklist for this process, which can serve as the critical criteria for completing it.

## 5. Application of Framework for a Case Study
This section will discuss how an illustrative case study can use the proposed framework to demonstrate applicability in real applications. The case study illustrates how AI can assist in data-driven strategy and planning to manage the hospital's resources effectively while ensuring the quality of care.

*Illustrative case study: Data-Driven Clinical Pathways in Hospital*
Effective data gathering, aggregation, and prediction of length of stay and cost-per-case during each patient admission is critical to plan the logistics of the workforce, equipment, medicines, and other products as needed by healthcare professionals to do their job, as well as mapping the processes and services required (Wang *et al.* 2018). Ultimately, such capability will help determine load and cost requirements to achieve cost efficiency and efficiency of care for the public health sector. AI-enabled decision support is essential to make rapid best-possible decisions when hospitals' resources are stretched during emergencies, major accidents, natural disasters, and epidemic diseases. For Intensive Care Units, accurate length of stay and cost-per-case calculation can help optimize resource allocation decisions and reduce the risk of readmission (Ponzoni *et al.* 2017, Lin *et al.* 2019). The solution cannot focus solely on technicalities. It is essential to work across the business model, strategy, planning, organization, and innovation management to enable smoother technology adoption to improve the existing processes and identify rooms for improvements across the whole procedure from admission to discharge.

| | | |
|---|---|---|
| Technological accessibility | <ul><li>The organization's existing internal infrastructure can support the AI solution, including people and data.</li><li>The organization commits to transform its operational processes to align with the AI solution and its required technologies, including data collection and management.</li></ul> | |
| AI-enabled business refinement | <ul><li>The initial solution can confirm the AI's feasibility and ability to serve a purpose, such as solving an identified problem or supporting business processes.</li><li>The organization's decision-makers understand the business case of the AI solution, in terms of its potential for delivering a return of investment, such as cost-saving, enhanced decisions, and improved business processes' efficiency.</li><li>Decision-makers understand the potential impacts on the existing human and technology infrastructure, and risks to the organization's reputation.</li></ul> | |
| Customer satisfaction | <ul><li>The AI solution brings direct benefits to end-users and improves customer experience.</li><li>Users can trust the AI solution and the organization that manages it due to clear manuals, documentations, and transparent risk assessments.</li><li>Users can use the AI-enabled services, including the time for learning (how to use), and buy the required equipment and fees to access it.</li></ul> | |
| Operational compliance | <ul><li>AI solution is compliant with regulations and policies and operates lawfully.</li><li>The impacts of operating and maintaining the AI solution can be managed by and compatible with the existing human and technology infrastructure.</li><li>AI solution aims to augment human abilities to support actions and decisions while ensuring the end-users' safety and wellbeing.</li><li>AI solution is equitable and particularly accessible for those who need the most.</li><li>AI solutions can be potentially used on a global scale without a significant shift in governance.</li></ul> | **Table IV.** Checklist for Plan and Design |
| Reliable operations | <ul><li>AI decisions and reasoning can be ascertained and trusted to support operation, and bias can be managed and reduced.</li><li>AI solution will function as intended and cannot be hacked and manipulated.</li></ul> | **Table V.** Checklist for development and deployment |
| Empowered employees | <ul><li>AI solutions can improve a human's ability to perform operational tasks and making decisions.</li></ul> | |
| | • AI-based decisions are consistent, reliable, fair, and aligned with social and cultural justice values in the local and global contexts.<br>• If applicable, the change management for transforming the workforce always ensures fair employment and labor practices. | |
| Engaged customers | • AI solution consistently meets performance expectations and increasingly more reliable to support or reduce manual work, make better decisions, or make services more enjoyable and personalized. | |
| Transformed products | • AI solutions can transform the existing norms via the new product(s) that people can enjoy and benefit from.<br>• AI solution (potentially) adheres with the existing ethical- and legal- theories or general principles. | |
| Continuous refinement | • AI solution can adapt and refine its model, improving its accuracy based on human feedback.<br>• AI solution can be continuously tested by research and development (R&D) and quality assurance process to ensure that the outcomes continue to meet the required tasks and performance expectations.<br>• AI solutions can continuously adapt to new operations and socio-economic requirements. | |
| Responsible business practice | • AI solutions can be managed by an end-to-end enterprise governance framework to be consistently accountable and transparent.<br>• AI's risks and scope of controls align with the organization's current and newly operationalized ethics. | |
| Individual (user) trust | • Human trust cannot be broken by being disadvantaged, such as being subjected to automated decisions, especially when such decisions have legsal ramifications.<br>• Users have the right to access all information needed to fully understand the AI's product and the test results appropriately before adoption.<br>• The policy and standard of practice are documented and transparent to ensure that users can trust the use of data and AI. | |
| Societal (community) trust | • An appropriate level of human supervisory is maintained to ensure that the AI solution continually aligns with human rights, social norms, and privacy regulations.<br>• Individual and collective AI use is transparent and auditable to ensure compliance with laws and fundamental rights and freedom. | Table VI. Checklist for operate, monitor, and scale-up |

Figure 3 illustrates the importance of data-driven decisions to assist in hospital planning and strategy. Data can support decisions across pre- and post-admission, including managing patient influx, predicting length of stay based on primary diagnosis, predicting the patient's pathway, length of stay, and the probability of readmission after discharge. As Figure 4 shows, the hospital staff can view a dashboard's predictive pathway and analyze the predicted length of stay (*LoS*) based on the possible complications. The descriptive analytics dashboard can help inform the types of information gathered by staff at an emergency department (*ED*) during patient presentation and triage to help predict LoS and decide the care plan. These include patient demographics data, disease types, and surgery occurrences. After the primary diagnosis, ED staff can use the predictive dashboard to inform the patient when discharge can be expected based on the predicted LOS, which may change depending on the possible complications.

As shown in Figure 5, the hospital staff's data will feed into the AI's descriptive and predictive analytics for analyzing the length of stay, readmission probability, and complications. The clinical experts need to provide feedback to fine-tune the AI model subsequently. The process from data collection to fine-tuning based on expert feedback should happen in real-time to support more accurate analytics.

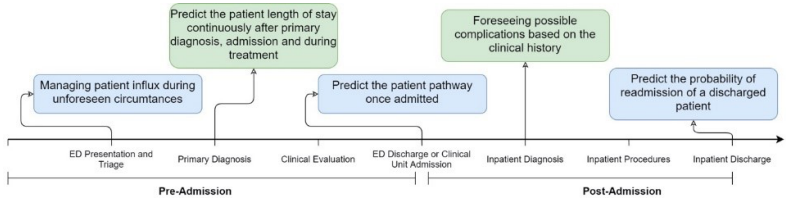

Figure 3. Strategies and Decisions in Hospital

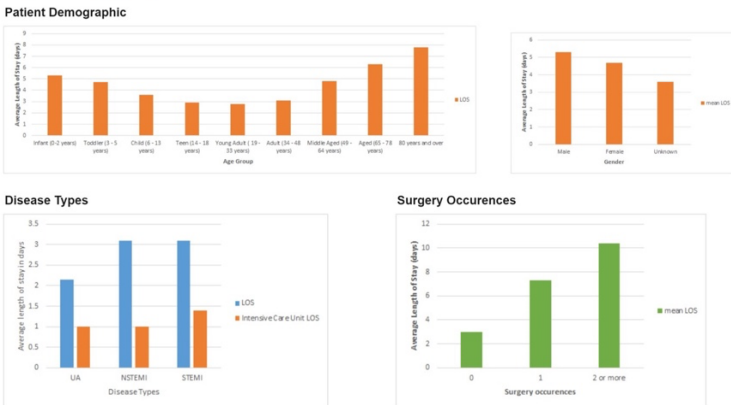

Figure 4. Descriptive (left) and Predictive(right) analytics dashboard of a patient's length of stay based on possible complications

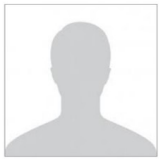
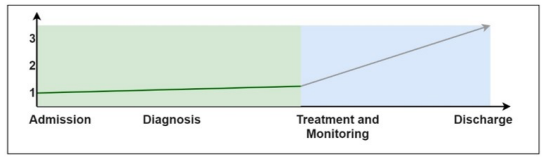
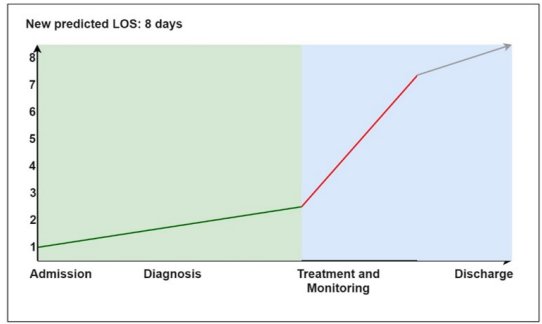

Figure 5. Descriptive (left) and Predictive(right) analytics workflow of a patient's length of stay based on possible complications

*Why the AI implementation requires Big Data and IoT adoption in this context*

There are two primary uses for predictive analytics. The first is clinical, to determine whether a patient is falling off the prescribed pathway and determine whether there is an opportunity to intervene and remedy this before significant complications occur. The second is administrative, which ensures costs are captured and that the clinical coding is accurate. The AI models for predictive analytics are only as good as the data used for training and fine-tuning. Big data about a patient needs to be collected across the admission, diagnosis, operation, treatment, and monitoring and discharge processes. Figure 6 illustrates the process of big data collection in a hospital workflow. At admission, the patient's electronic health record (*EHR*) can form the basis for initial patient information, which will be augmented with the clinical history and appointment regularity. At the diagnosis stage, disease type and patient risk type determines the need for various treatments, including potential surgery. When surgery occurs, the incidental finding helps to determine if a patient needs to be sent to another department for further diagnosis or sent for monitoring based on whether the patient is high-risk or not. During the treatment and monitoring stage, the system needs to predict and potentially prevent significant complications resulting in morbidity or mortality. Moreover, unless there is a nosocomial infection, the patient's ongoing status data will recommend discharge timing. At discharge, clinicians enter vital information, including data for the patient administration system to capture the patient's journey and experience, and the discharge disposition record for supporting follow-up by primary care providers.

The amount of data collected depends on the patient's clinical pathway. Therefore, IoT can connect data entry points and the hospital equipment to minimize manual data entry. For example, patients ' vital data like heart rate, blood pressure, and oxygenation can be collected directly from the sensor devices. Real-time data collection is especially useful for high-risk patients and during critical stages in intensive care. Moreover, the AI model needs to work with the hospital's existing information systems so that when new data is entered into the database, the AI system is immediately notified to perform real-time calculation and prediction of the patient's LoS and pathway.

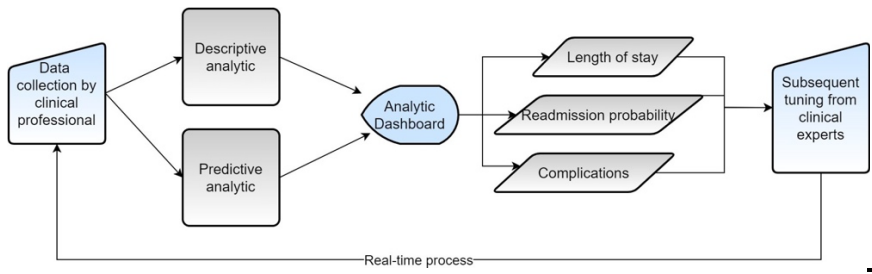

**Figure 6.** The need for capturing big data across a clinical pathway

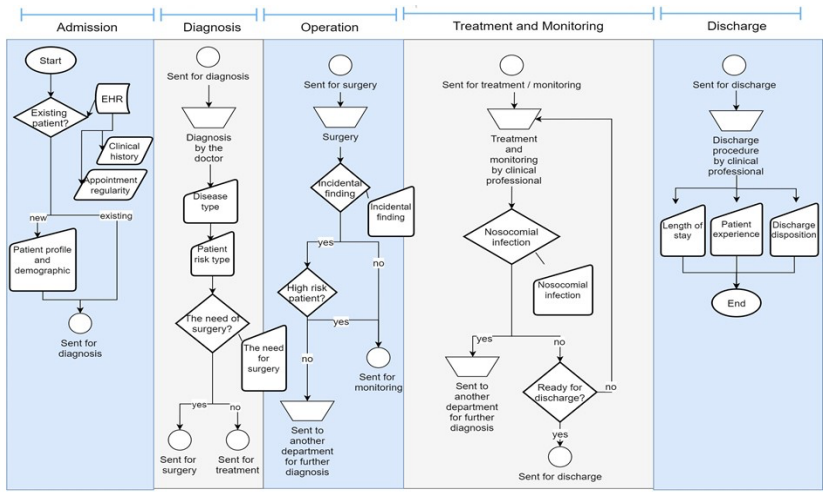

**Figure 7.** The need for capturing big data across a clinical pathway

*Importance of initial trust and maintaining it across implementation process*
The initial trust is important to gain internal support for adopting the AI and data collection process. The ongoing trust is key to ensure that the AI decisions are transparent and explainable and consistently comply with the policies, laws, and regulations, as well as show consistency in terms of performance and reliability.

Across the three stages of AI implementation, forming and maintaining trust is critical to:
- Design and implement an effective and secure AI-enabled information system to support user interactions and assist with decisions and strategies.
- Establish a "live" test environment for the system's infrastructure (technology and people) that will be used, researched, and documented alongside the current AI system deployment. The test environment will support a more accurate evaluation of the system's usefulness while involving all stakeholders to test the business model, policies, and procedures.
- Develop and recommend strategies to enable rapid scale-up and adoption across states and nationally. The adoption roadmap needs to recognize the variations across the stakeholders, from clinicians and departments in the hospital to the Public health system. Each hospital is different in how it functions, each patient cohort is different, and treatment options vary from institution to institution. Therefore, the test environment will need to be re-tested for any scale-up to recognize the need to adapt the AI system to the available data and the system functionalities.

An existing roadmap for deploying effective machine learning systems in health care also emphasizes engaging stakeholders early in the process so that issues stemming from the complexity of adopting AI in practice can be avoided (Wiens *et al.* 2019). Our three-process iterative framework RAIIF can be aligned with their roadmap's checklist:
- Choosing the right problems: clinical relevance, appropriate data, collaborators, and definition of success
- Developing a useful solution: data provenance and ground truth to develop AI models.
- Considering ethical implications: ethicist engagement and bias correction.
- Rigorous evaluation and thoughtful reporting: model use, sensical predictions, shared model/code, failure modes
- Deploying responsibly: prospective performance, clinical trial, and safety monitoring
- Making it to the market: medical device, model updates

However, the processes' sequencing should be shifted to focus on the initial trust formation and the subsequent ongoing trust maintenance via iterative research and monitoring of the AI implementation. Choosing the right problems and strategy to deliver to the market (including the required medical device investment and model updates) must be placed at the start of planning and designing the ecosystem of an AI solution. Developing useful solutions, rigorous evaluation, and thoughtful reporting should focus on iterative development and deployment. Finally, the underlying principle from the planning until the large-scale deployment stems from ethical implications and responsible deployment responsibly. Therefore, AI performance management, clinical trial, and safety monitoring need to be co-designed, tested, and evaluated with ethical implications and a return of investment into consideration throughout the whole process.

*Reflections from practitioners on the framework*
Using the case study, we examine and reflect on the applicability of the proposed framework from practitioners' standpoint. Each practitioner reflects on their specific roles on each of the key aspects of the framework:
- Siiri Hatakka: project manager and business development of Everledger
- Brent Richards: director of ICU at a Hospital
- Damian Green: chief information officer (CIO) of eHealth of a State government

Their considerations serve as practitioners' mediating expertise to capture the nuances of real-world challenges and provide further details that complement the checklists in Tables IV, V, and VI.

*The importance of trust and the co-creation process*
Transparency about the state of the technology is important to move from ambition to execution. "*In my experience, the complexity and the maturity of the technology solution are often miss communicated in the initial trust formation stage. There are many reasons for this, but the bias is often a combination of the identified potential of the AI solution and the motivation of the technology provider to sell the solution. Moreover, this contributes greatly to the gap between the ambition and execution for adopting AI*" (S. Hatakka).

A shared and agreed vision on a problem worth solving is crucial to motivate trust formation and the co-creation process. "*Once the problem worth solving is clear, the triumvirate of a patient, clinician, and executive are more likely to work to improve it, accept changes to workflow, and accept the continuing adaptation*." (B. Richards).

AI technology in healthcare is generally not mainstream yet, so there is no precedence in realizing the benefits and risks of adoption. An inflated business case or overestimation of the AI's capability may diminish the early trust in the technology. "*The hype and expectation baggage of new and untested technology is determined by the business's journey to understand the technology's capabilities accurately in practicality. The challenge is to overcome the risks from short-term expectations over the return-of-investment*". (D. Green).

*The key considerations for plan and designing the ecosystem*
The initial business case for AI Adoption needs to incorporate an accurate timeline and budget implications to establish and sustain trust. *"When we consider technology and its role in enabling digital transformation, it is not only about the suitability of the solution, but often it is about the combination of the complexity and maturity of the right solution that significantly impacts the timeline of the adoption roadmap. It is important from the industry perspective to talk about technology adoption timelines because the deployment timeline often has significant budget implications. The mistakes in planning the solution development timeline degrade the trust for the technology for implementing the AI solution full scale."* (S. Hatakka).

A key consideration for planning to implement AI into a hospital ecosystem is understanding that there is a system on systems within the technology, people, and processes. The AI solution must work with the current ecosystem, including legacy processes and information systems, while minimizing data collection and processing complexity. *"Clinicians are already very busy, and thus any new system must improve patient care and outcomes by improving the use of current data without requiring new data entry, decreasing rather than increasing workload."* (B. Richards).

Contexts and timing of AI use are critical decisions, as the organization needs to chart the pathway for implementing the technology into the existing system and processes. *"To plan right, we should think big but start small while taking into account the organization's AI maturity. The implementation framework needs to consider the ordering of the processes to adopt, expand, and transform the technology depending on the state of the technology and the organization's readiness."* (D. Green).

*The key considerations for iterative development and deployment*
In general, keep iterations at a minimum to keep everyone motivated. "*As a project manager, I always have to think about time, resources, and money… What I find the trouble is that with solutions that are custom-made or very new, it is hard to achieve the needed accuracy with as few iterations as possible to keep everybody motivated. AI as a concept is highly complex and impacts several aspects of business, such as operations, data, IoT, investment, business models, etc. Therefore, multiple iterations of AI development - involving machine learning and testing – often become unattractive for the business due to high cost and complexity. This is commonly perceived as a high risk for business, and only the early adopters are very willing to execute projects requiring multiple iterations. It also depends on the company, as some businesses are very agile and comfortable with a high*

*number of iterations. However, some still prefer a waterfall type of planning and execution due to its existing culture and organizational structure*." (S. Hatakka).

Trust in AI technology for medical applications needs to consider biases in the AI models and the data used for training. The AI solution needs to go through the same rigors as any other clinical intervention, and it needs to be studied within the target population properly. All stakeholders (involved in the implementation) would benefit from using the same language to understand and appreciate each other's considerations during the iterative process. Executive support is critical to support the cost and manage the timeline and return of investment. *"Successful implementation of AI solution requires maintaining the hospital staff's trust and confidence in the technology's capability 'to maintain excellence in clinical care.*" (B. Richards)

The design, development, and evaluation stage need to be agile, iterative, and not rigid. Some AI implementation processes should be conducted concurrently and non-linear, instead of a waterfall-like software development method. The engagement and partnership with end-users or consumers are critical. Therefore we need to develop mechanisms to obtain feedback and establish a social contract (or digital charter) with the consumers. *"Health consumers can trust tech-enabled healthcare solutions if they can be assured that the technology is transparent and their data is used appropriately"* (D. Green).

*The key considerations for operating, monitoring, and scaling*
The project manager needs to plan and execute within the timeline and budget to enable people to continue working on implementing a new AI solution. *"Eventually, the industry can adopt anything that makes sense, but for successful technology implementation, the managers have to plan and budget for the business model and technology transformations. The timeline of adopting the solution has significant implications for people's motivation to continue working on implementing the new technology."* (S. Hattaka).

The AI solution needs to fit into the organizational workflow and can be supported by the IT team to enable an ongoing operation of AI solution. Innovation within a hospital context is managed by a transformation group and directed by a director of innovation. At this management level, the AI solution needs to support clinicians to work on specific targets. *"Prediction of patient's pathway and LoS is key for predicting when the hospital will be full and the number of patients admitted while maintaining optimum quality of care. Sustained trust in the AI solution to achieve these goals will depend on the AI solution's ability to recognize the variants in care and continuously monitor the AI models' performance and biases."* (B. Richards)

AI drives data flows and system architecture, therefore, data privacy and governance are the critical factors for maintaining trust within the long-term use of AI. "*Strategy for AI adoption needs to ensure data protection and that it can work with the existing IT system. A CIO's role is to ensure that information is always accurate and protected. There needs to be transparency on how data is generated and used by AI appropriately.*" (D. Green).

## 6. Conclusions and Future Work
This paper proposes a new theoretical framework for responsible AI implementation in businesses based on a critical review of academics-led and practitioners-led literature. The key contribution is emphasizing a synergistic business-technology approach underpinned by a human-centered design framework and agile co-creation process. Given the significant gap between expectation and successful AI implementation in businesses, the framework is fundamentally driven by the formation and maintenance of trust by involving all stakeholders throughout the project. Once a shared vision and viable business case for gradual implementation of an AI solution are achieved, the business leaders and executive decision-makers need to commit the resource, time, and budget. The leadership will ensure the technology is designed, developed, and deployed in conjunction with people and not in isolation. Each process in the framework should be iterative (not linear) and guided by a set of checklists underpinned by the requirements of responsible AI in the era of the Internet of

Things and big data. The paper presents an illustrative case study and mediating practitioners' expertise to reflect on the key considerations for AI-enabled hospital planning and decision-making. The aim is to demonstrate that the proposed framework is practical and applicable to real-life applications.

For future work, the framework should adapt and co-evolve within a project for implementing responsible AI in businesses at various maturity stages. Based on the proposed iterative processes and the checklists to ensure that responsible AI is achieved at every stage, all stakeholders will participate in the design, development, and evaluation of the AI solution. The framework can be extended for assessing its benefits over a long-term longitudinal observation of the AI implementation projects using qualitative (e.g., interviews, focus groups, and documentation from the co-creation processes) and quantitative (e.g., surveys) data collection and analysis from all stakeholders at the different stages of a short-term. The framework can be further adjusted to suit different industry sectors and AI maturity stages and fine-tuned for different organizational characteristics. The adjustments will focus on improving the practicality, comprehensiveness, and effectiveness of the general and tailored approach to applying the proposed framework in various case studies.